\documentclass[twoside,final,1p,times]{elsarticle}
\usepackage{amsmath}
\usepackage{amsfonts}
\usepackage{amssymb}
\usepackage{graphicx}
\usepackage{multirow}
\usepackage{url}
\usepackage{fancyhdr}
\usepackage{color}
\usepackage{booktabs}
\usepackage{tabularx}
\usepackage{array}
\newcolumntype{Y}{>{\raggedright\arraybackslash}X}
\usepackage{mathtools}
\usepackage{algorithmic}
\usepackage{algorithm}
\usepackage{textcomp}
\usepackage{hyperref}
\usepackage[section]{placeins}
\makeatletter
\let\ps@pprintTitle\ps@plain
\makeatother
\hypersetup{pdftitle={S2TDR: Structured Multimodal Reconstruction for Sticker-Aware Dialogue Understanding},pdfauthor={Yuanchen Shi and Fang Kong}}

\begin{document}
\begin{frontmatter}

\title{S2TDR: Structured Multimodal Reconstruction for Sticker-Aware Dialogue Understanding}

\author[1]{Yuanchen Shi}
\ead{20227927002@stu.suda.edu.cn}
\author[1]{Fang Kong\corref{cor1}}
\ead{kongfang@suda.edu.cn}
\cortext[cor1]{Corresponding author.}
\address[1]{Natural Language Processing Laboratory, School of Computer Science and Technology, Soochow University, Suzhou, China}

\begin{abstract}
Stickers are widely used in social media dialogues to express affect, stance, and intent beyond literal text. Their meanings are often non-literal and context-dependent, making them difficult to handle with generic visual fusion or standalone captioning. We propose Sticker-to-Text Dialogue Reconstruction (S2TDR), a structured multimodal reconstruction framework that converts sticker turns into contextually appropriate textual utterances. S2TDR decomposes sticker interpretation into visual cue extraction, affective-intent inference, and context-aware dialogue reconstruction, using OCR text only as an auxiliary cue when available. The reconstructed dialogue can then be consumed by existing text-based dialogue models. Experiments on sentiment analysis, intent recognition, and dialogue summarization show that S2TDR generally improves multiple downstream models over direct sticker integration methods. Further analyses on component ablations, OCR-empty stickers, and reconstruction quality indicate that structured reconstruction provides an interpretable and effective way to integrate sticker semantics into dialogue understanding.
\end{abstract}


\end{frontmatter}

\section{Introduction}
\label{sec:intro}

Stickers are pervasive in social media dialogues, where they often express stance, affect, and intent beyond literal text. Unlike ordinary images, stickers frequently function as pragmatic acts: their meanings are stylized, non-literal, and strongly shaped by the surrounding turns. This makes sticker-aware dialogue understanding challenging. A model must infer not only what a sticker depicts, but also what role it plays in the conversation. Existing studies have introduced sticker-related tasks and datasets, including retrieval \cite{Wang_Du_Liang_Bai_Yang_Wang_Wong_Xu_2025}, generation \cite{yuan2025vsd2mlargescalevisionlanguagesticker}, multimodal sentiment analysis \cite{ge-etal-2022-towards}, and emotion recognition \cite{liu2022ser30k}. However, effective and interpretable ways to integrate sticker semantics into downstream dialogue models remain limited.

\begin{figure}[!t]
\centering
\includegraphics[scale=0.8]{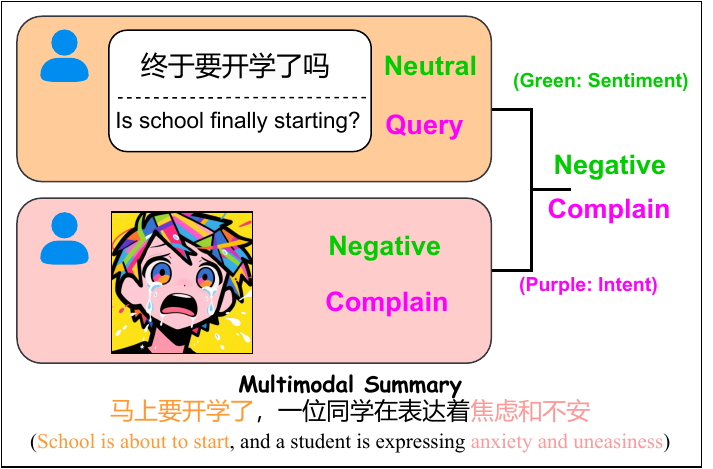}
\caption{A dialogue example with text and a sticker.}
\label{fig:example}
\end{figure}

Figure~\ref{fig:example} illustrates the problem. The crying-character sticker does not merely add visual content; it conveys anxiety and unease, which affects both sentiment and intent interpretation. For summarization, the surrounding text provides the factual event, while the sticker contributes the implicit emotional state. This suggests that sticker understanding should be context-conditioned and transformed into a representation that downstream dialogue models can reliably use.

To address this challenge, we propose Sticker-to-Text Dialogue Reconstruction (S2TDR), a structured reconstruction framework that converts sticker turns into contextually appropriate textual utterances. Given a multimodal dialogue, S2TDR interprets each sticker through three components: visual cue extraction, affective-intent inference, and context-aware dialogue reconstruction. OCR text is used only when available; for OCR-empty stickers, the reconstruction relies on visual cues and dialogue context. The resulting dialogue is fully text-based and can be directly used by existing text encoders, sequence-to-sequence models, or large language models.

We evaluate S2TDR on two sticker-aware dialogue benchmarks covering sentiment analysis, intent recognition, and dialogue summarization. Results show that structured reconstruction brings broad improvements over direct sticker integration methods such as emotion-intent replacement and description appending. To examine why the framework works and where it may fail, we further conduct component ablations, OCR-empty sticker analysis, reconstruction-quality evaluation, and qualitative error analysis.

Our contributions are summarized as follows:
\begin{itemize}
    \item We formulate sticker-to-text dialogue reconstruction as an interpretable intermediate representation for sticker-aware dialogue understanding.
    \item We propose S2TDR, a structured multimodal reconstruction framework that converts sticker semantics into context-conditioned textual utterances.
    \item We show that S2TDR improves multiple downstream models on sentiment analysis, intent recognition, and dialogue summarization.
    \item We provide diagnostic analyses on component contributions, OCR-empty stickers, reconstruction quality, and typical failure cases.
\end{itemize}

\section{Related Work}
\label{sec:related}

Sticker-related multimodal understanding has received increasing attention in recent years. Existing work has introduced datasets and tasks for sticker-aware sentiment analysis, emotion recognition, sticker tagging, retrieval, generation, and empathetic dialogue modeling \cite{ge-etal-2022-towards,liu2022ser30k,wang2024towards,Wang_Du_Liang_Bai_Yang_Wang_Wong_Xu_2025,zhang2024stickerconv_acl,yuan2025vsd2m,chee2025gse_triples,chee2025stickerqueries}. These studies show that stickers carry rich affective and semantic information, but most of them focus on recognition, retrieval, or generation. The context-dependent pragmatic role of stickers in downstream dialogue understanding remains less explored.

Our work is also related to multimodal affect modeling in dialogue. Prior studies on multimodal emotion recognition and vision-grounded dialogue modeling typically fuse textual, visual, or acoustic features through attention-based or joint encoders \cite{wu2025merc_survey,yu-etal-2021-vision}. Emoji-aware and sarcasm-oriented studies further show that visual paralinguistic symbols can substantially affect pragmatic interpretation \cite{chauhan2022seemoji_mustard}. However, stickers differ from ordinary images: they are stylized, non-literal, and often used as conversational acts. Direct feature fusion or standalone captioning may therefore miss how a sticker functions within the dialogue context.

Finally, staged reasoning with large multimodal models provides a practical way to decompose complex multimodal inference. Existing studies have explored multimodal chain-of-thought reasoning, perception--decision decomposition, and intermediate scene-graph-style reasoning for visual-language tasks \cite{zhang2023multimodalcot,gao2024cantor,mitra2024ccot,chainofx2025survey}. In contrast to proposing a general prompting algorithm, we use staged reasoning as a task-specific reconstruction mechanism: S2TDR converts sticker turns into context-conditioned textual utterances, making sticker semantics directly usable by downstream dialogue models.

\begin{figure}[!t]
\centering
\includegraphics[scale=0.8]{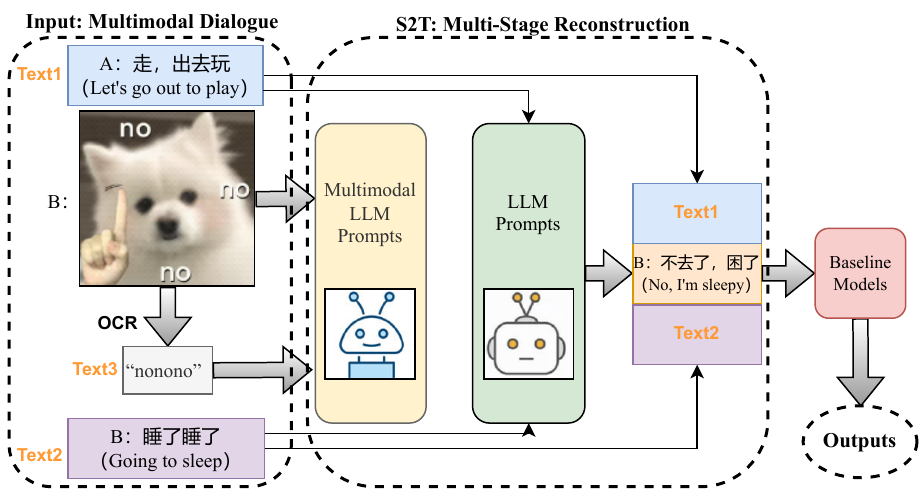}
\caption{Overview of S2TDR.}
\label{fig:method}
\end{figure}

\begin{figure}[t]
\centering
\includegraphics[width=0.95\linewidth]{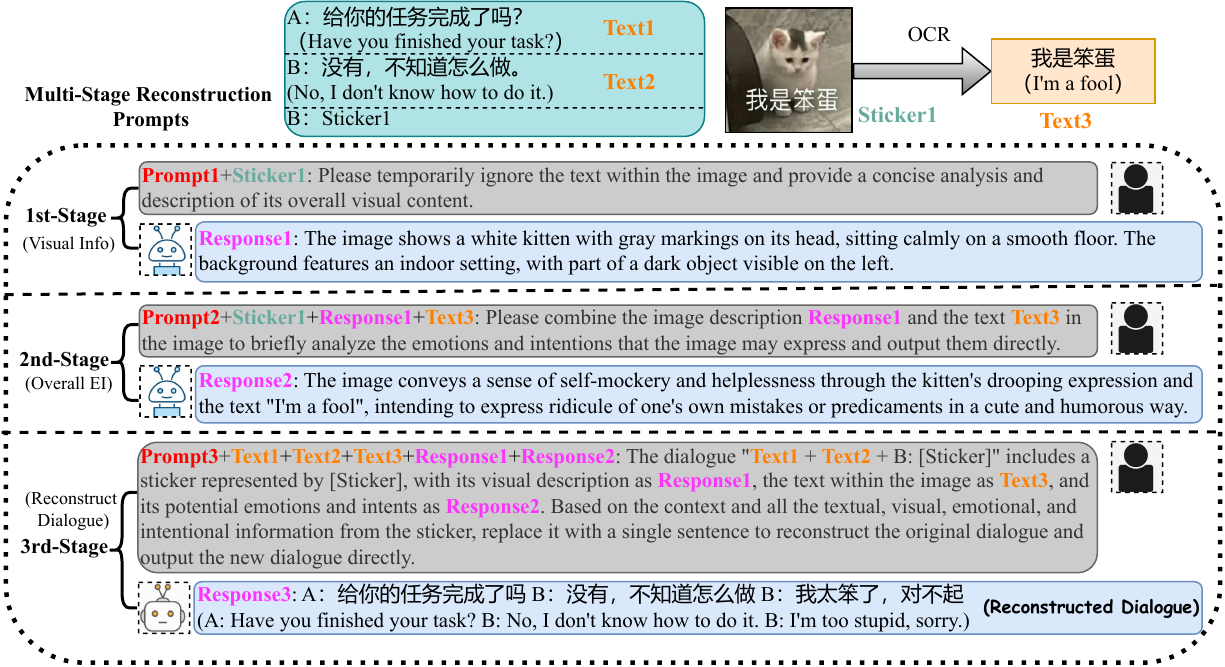}
\caption{Staged sticker-to-text reconstruction.}
\label{fig:prompt}
\end{figure}

\section{S2TDR Framework}
\label{sec:framework}

\subsection{Task Formulation}
\label{sec:task}

Given a multimodal dialogue $D=\{x_i\}_{i=1}^{n}$ containing both textual utterances and sticker turns, S2TDR produces a fully text-based dialogue $D'$ that preserves the affective and intentional meaning conveyed by stickers. The reconstructed dialogue is then used by a task-specific downstream model:
\[
D' = \mathrm{S2TDR}(D), \qquad y = f(D'),
\]
where $f(\cdot)$ denotes a model for sentiment analysis, intent recognition, or dialogue summarization. This formulation treats reconstruction as an intermediate representation: S2TDR does not replace downstream models, but converts sticker semantics into a textual form that existing dialogue models can directly consume.

\subsection{Framework Overview}
\label{sec:overview}

Figure~\ref{fig:method} shows the overall framework. S2TDR converts each sticker turn into a contextually appropriate textual utterance through three components: visual cue extraction, affective-intent inference, and context-aware dialogue reconstruction. OCR text, when available, is used as an auxiliary cue rather than a required input. For OCR-empty stickers, the framework relies on visual cues and dialogue context.

The reconstructed dialogue keeps the original conversational structure while replacing sticker images with textual utterances. This makes the output interpretable and compatible with text encoders, sequence-to-sequence models, and instruction-tuned language models.

\begin{figure}[t]
\centering
\includegraphics[width=0.7\linewidth]{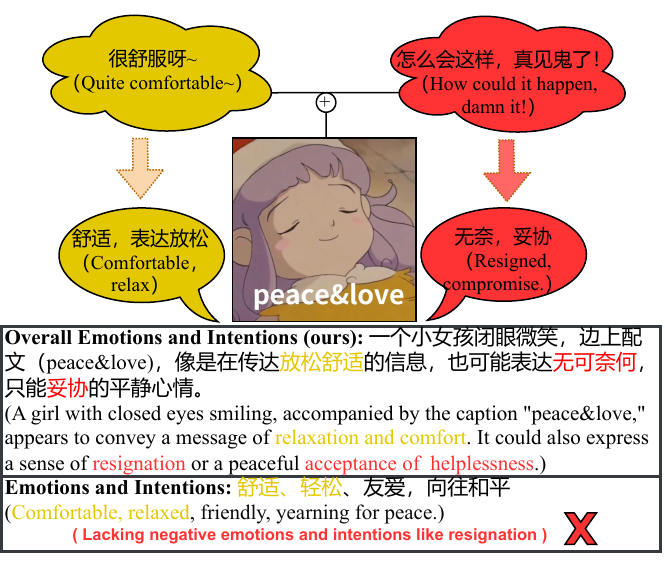}
\caption{Context helps infer sticker emotion and intent.}
\label{fig:OEI}
\end{figure}

\subsection{Structured Sticker-to-Text Reconstruction}
\label{sec:multicot}

S2TDR uses a structured decomposition rather than rewriting a sticker in one step. Sticker interpretation involves three levels of information: what the sticker depicts, what affective or intentional meaning it conveys, and how that meaning should be expressed in the current dialogue. Separating these levels makes the reconstruction more interpretable and easier to ablate.

Figure~\ref{fig:prompt} illustrates the reconstruction flow. S2TDR first extracts visual cues from the sticker, then infers the sticker's overall emotion and intent by combining visual information with OCR text when available, and finally rewrites the sticker as a concise utterance conditioned on the dialogue context.

\paragraph{Visual cue extraction}
For each sticker image, a multimodal language model extracts salient visual cues such as expression, posture, objects, and style. PaddleOCR~\cite{du2020pp} extracts embedded text when available, but OCR is treated only as an auxiliary cue. For OCR-empty stickers, visual cues provide the main perceptual evidence.

\paragraph{Affective-intent inference}
The visual description and OCR text are used to infer the sticker's overall emotion and intent, such as self-mockery, sarcasm, complaint, encouragement, or compromise. This component summarizes the pragmatic role of the sticker before it is rewritten as dialogue.

\paragraph{Context-aware reconstruction}
S2TDR conditions on the dialogue context, OCR text, visual description, and inferred emotion-intent information to generate a single utterance that replaces the sticker turn. The generated utterance is constrained to match the original speaker and remain concise, reducing unrelated expansion.

Figure~\ref{fig:OEI} shows why context is necessary. A sticker that appears relaxed in isolation may express helplessness or compromise when paired with a negative utterance. S2TDR therefore reconstructs stickers according to their dialogue function rather than treating them as standalone images.

\subsection{Context Construction}
\label{sec:context}

Each dialogue turn is represented as $x_i=(s_i,c_i,k_i)$, where $s_i$ is the speaker, $c_i$ is the content, and $k_i \in \{\mathrm{text}, \mathrm{sticker}\}$. Text turns are kept unchanged. For a sticker turn, S2TDR constructs a local context and maps the sticker into a textual replacement:
\[
c_i' = \mathrm{S2T}(c_i, \mathrm{context}(x_i)).
\]

The context is constructed sequentially. Preceding sticker turns are replaced with their reconstructed text, allowing later reconstructions to use an updated dialogue history. Succeeding turns are included only when they are textual, avoiding circular dependence on future sticker reconstructions. The final output is a text-only dialogue $D'=\{(s_i,c_i')\}_{i=1}^{n}$, which is fed into the downstream model.

\section{Experiments}
\label{sec:experiments}

\subsection{Datasets and Tasks}
\label{sec:datasets}

We evaluate S2TDR on two sticker-aware dialogue benchmarks. MSAIRS is used for multimodal sentiment analysis and intent recognition \cite{shi2025impact}, while MCDSCS is used for multimodal dialogue summarization \cite{shi2024integrating}. These datasets cover both classification-style understanding and generation-style dialogue modeling.

For MSAIRS, we report accuracy and macro F1 for sentiment analysis and intent recognition. For MCDSCS, we follow prior work and report ROUGE-1, ROUGE-2, ROUGE-L \cite{lin-2004-rouge}, BLEU \cite{graham-2015-evaluating}, and METEOR \cite{banerjee-lavie-2005-meteor}. To examine the role of readable sticker text, we report OCR/text coverage in both datasets. As shown in Table~\ref{tab:ocr-coverage}, OCR text is common but not universal: over half of MSAIRS samples have empty sticker text, and MCDSCS also contains many OCR-empty stickers. This supports our treatment of OCR as an auxiliary cue rather than a required input.

\begin{table}[t]
\centering
\small
\caption{OCR/text coverage.}
\label{tab:ocr-coverage}
\begin{tabular}{llrrrr}
\toprule
Dataset & Unit & Total & Non-empty & Empty & Empty ratio \\
\midrule
MSAIRS & sample/sticker & 3,118 & 1,482 & 1,636 & 52.47\% \\
MCDSCS & unique sticker & 14,356 & 9,719 & 4,637 & 32.30\% \\
MCDSCS & dialogue sample & 5,527 & 4,951 & 576 & 10.42\% \\
\bottomrule
\end{tabular}

\vspace{2pt}
\footnotesize
\emph{Note:} For MCDSCS dialogue samples, ``Non-empty'' means that at least one sticker has OCR text, while ``Empty'' means that all stickers are OCR-empty.
\end{table}

\subsection{Implementation Details}
\label{sec:setup}

In the default setting, visual cue extraction is implemented with LLaVA-v1.6-7B \cite{li2024llava}, while affective-intent inference and dialogue reconstruction use Qwen2.5-7B-Instruct \cite{wang2024qwen2}. PaddleOCR \cite{du2020pp} is used to extract embedded sticker text when available. For OCR-empty stickers, S2TDR relies on visual descriptions and dialogue context.

The prompt templates are fixed across datasets. The first component produces a concise visual description, the second summarizes the sticker's emotion and intent, and the third generates a single utterance to replace the sticker turn. We use low-temperature decoding and constrain the reconstructed utterance length to reduce irrelevant expansion. The prompt flow is illustrated in Figure~\ref{fig:prompt}; the same templates are used across datasets unless otherwise specified.

\begin{table}[t]
\centering
\caption{Results on MSAIRS.}
\label{tab:msairs}
\begingroup
\footnotesize
\setlength{\tabcolsep}{3.5pt}
\renewcommand{\arraystretch}{0.88}
\begin{tabular*}{0.90\linewidth}{@{\extracolsep{\fill}}lcccc@{}}
\toprule
\multirow{2}{*}{Model} & \multicolumn{2}{c}{Sentiment} & \multicolumn{2}{c}{Intent} \\
\cmidrule(lr){2-3}\cmidrule(l){4-5}
& Acc. & F1 & Acc. & F1 \\
\midrule
\multicolumn{5}{c}{\textit{Fine-tuned models}} \\
\midrule
BERT-base \cite{devlin2019bert} & 65.05 & 64.48 & 61.94 & 61.84 \\
BERT-EI & 65.45 & 65.09 & 66.18 & 66.30 \\
BERT-ITEI & 66.90 & 67.51 & 65.45 & 65.21 \\
BERT-S2TDR & \textbf{69.09} & \textbf{68.11} & \textbf{66.90} & \textbf{66.78} \\
\midrule
RoBERTa-base \cite{liu2019roberta} & 66.78 & 66.55 & 65.74 & 65.40 \\
RoBERTa-EI & 66.91 & 66.30 & 68.36 & 68.35 \\
RoBERTa-ITEI & 67.27 & 66.91 & 65.65 & 65.34 \\
RoBERTa-S2TDR & \textbf{68.82} & \textbf{68.46} & \textbf{68.77} & \textbf{68.60} \\
\midrule
MMSAIR \cite{shi2025impact} & 70.58 & 70.91 & 72.31 & 72.29 \\
MMSAIR-EI & 69.32 & 70.17 & 73.51 & 73.74 \\
MMSAIR-ITEI & 70.49 & 70.62 & 71.89 & 72.05 \\
MMSAIR-S2TDR & \textbf{71.49} & \textbf{71.28} & \textbf{74.45} & \textbf{74.96} \\
\midrule
w/o visual cues & 71.05 & 70.77 & 73.58 & 73.61 \\
w/o affective-intent inference & 70.78 & 70.75 & 72.43 & 72.62 \\
w/o dialogue reconstruction & 70.11 & 69.84 & 71.23 & 71.03 \\
\midrule
\multicolumn{5}{c}{\textit{Zero-shot LLMs}} \\
\midrule
GLM4 \cite{glm2024chatglmfamilylargelanguage} & 55.27 & 55.42 & 37.81 & 37.48 \\
GLM4-EI & 63.27 & \textbf{63.79} & 41.82 & 42.11 \\
GLM4-ITEI & 54.54 & 54.09 & 36.13 & 35.98 \\
GLM4-S2TDR & \textbf{63.46} & 63.51 & \textbf{49.09} & \textbf{48.92} \\
\midrule
Qwen2.5-7B \cite{wang2024qwen2} & 29.32 & 29.17 & 29.27 & 28.94 \\
Qwen2.5-EI & 30.23 & 30.05 & 30.18 & 30.56 \\
Qwen2.5-ITEI & 28.59 & 29.02 & 28.54 & 28.73 \\
Qwen2.5-S2TDR & \textbf{32.60} & \textbf{32.30} & \textbf{32.54} & \textbf{32.47} \\
\midrule
\multicolumn{5}{c}{\textit{Direct multimodal baselines}} \\
\midrule
mBERT \cite{yu2019adapting} & 68.86 & 71.89 & 65.05 & 58.40 \\
PMF \cite{li2023efficient} & 65.95 & 62.27 & 52.60 & 47.01 \\
CSMSA \cite{ge-etal-2022-towards} & 63.96 & 61.87 & 57.48 & 55.07 \\
Yi-VL-6B \cite{young2024yi} & 37.82 & 32.50 & 5.10 & 6.09 \\
LLaVA-v1.6-7B \cite{li2024llava} & 38.18 & 34.15 & 14.55 & 16.49 \\
Qwen2.5-VL-7B \cite{wang2024qwen2} & 35.64 & 31.01 & 5.82 & 6.84 \\
GPT-4o & 47.59 & 45.23 & 32.56 & 32.17 \\
\bottomrule
\end{tabular*}
\endgroup
\end{table}

\subsection{Baselines and Sticker Integration Methods}
\label{sec:baselines}

We compare S2TDR with two groups of methods. For downstream models, MSAIRS includes fine-tuned text encoders, MMSAIR, and zero-shot LLMs, while MCDSCS includes encoder--decoder summarization models, an instruction-tuned LLM, and representative multimodal baselines.

For sticker integration, we compare four settings: the original input without explicit reconstruction; EI \cite{shi2024integrating}, which replaces stickers with emotion-intent labels; ITEI \cite{shi2025msa}, which appends prompt-derived sticker descriptions; and S2TDR, which rewrites sticker turns into context-conditioned textual utterances. This comparison isolates how sticker information is incorporated from the choice of downstream architecture.

\begin{table}[t]
\centering
\caption{Results on MCDSCS.}
\label{tab:mcdscs}
\begingroup
\footnotesize
\setlength{\tabcolsep}{3.2pt}
\renewcommand{\arraystretch}{0.88}
\begin{tabular*}{0.95\linewidth}{@{\extracolsep{\fill}}lccccc@{}}
\toprule
Model & R1 & R2 & RL & BL & MT \\
\midrule
\multicolumn{6}{c}{\textit{Fine-tuned models}} \\
\midrule
BART-large \cite{lewis-etal-2020-bart} & 44.07 & 21.17 & 38.11 & 14.60 & 29.74 \\
BART-EI & 45.67 & 22.38 & 36.86 & 15.09 & 30.22 \\
BART-ITEI & 46.98 & 23.52 & \textbf{39.83} & 15.25 & 30.44 \\
BART-S2TDR & \textbf{47.24} & \textbf{23.71} & 39.59 & \textbf{15.37} & \textbf{30.59} \\
\midrule
T5-base \cite{raffel2020exploring} & 31.82 & 13.49 & 26.91 & 9.14 & 16.24 \\
T5-EI & 31.99 & 13.89 & 27.22 & \textbf{9.73} & 16.86 \\
T5-ITEI & 34.38 & \textbf{14.93} & 28.91 & 9.48 & 17.26 \\
T5-S2TDR & \textbf{36.86} & 14.49 & \textbf{31.42} & 9.58 & \textbf{17.47} \\
\midrule
Pegasus-large \cite{zhang2020pegasus} & 43.08 & 18.35 & 36.08 & 11.79 & 26.87 \\
Pegasus-EI & 44.24 & 19.80 & 37.26 & 12.58 & 28.31 \\
Pegasus-ITEI & 44.70 & 20.10 & 38.00 & 12.50 & 27.79 \\
Pegasus-S2TDR & \textbf{47.36} & \textbf{24.25} & \textbf{41.62} & \textbf{15.00} & \textbf{31.13} \\
\midrule
LLaMA2-7B \cite{touvron2023llama} & 52.31 & 25.00 & 42.42 & 13.69 & 39.21 \\
LLaMA2-EI & 55.74 & 36.07 & 46.87 & 15.55 & \textbf{44.45} \\
LLaMA2-ITEI & 59.15 & 35.90 & 47.89 & 18.69 & 42.55 \\
LLaMA2-S2TDR & \textbf{59.45} & \textbf{36.83} & \textbf{48.87} & \textbf{19.13} & 42.15 \\
\midrule
w/o visual cues & 59.12 & 36.17 & 48.11 & 18.43 & 42.01 \\
w/o affective-intent inference & 57.92 & 35.91 & 47.73 & 17.49 & 41.86 \\
w/o dialogue reconstruction & 54.69 & 36.01 & 46.56 & 15.98 & 41.23 \\
\midrule
\multicolumn{6}{c}{\textit{Zero-shot and multimodal baselines}} \\
\midrule
GPT-3.5 & 38.26 & 15.32 & 31.69 & 9.48 & 20.97 \\
GPT-3.5-EI & 41.57 & 18.85 & 35.42 & 10.03 & 21.88 \\
GPT-3.5-ITEI & 43.03 & 18.94 & 36.11 & 11.38 & 22.79 \\
GPT-3.5-S2TDR & \textbf{43.59} & \textbf{19.18} & \textbf{36.87} & \textbf{13.93} & \textbf{23.55} \\
\midrule
VG-BART \cite{yu-etal-2021-vision} & 45.58 & 21.90 & 35.23 & 14.09 & 28.44 \\
VG-T5 \cite{yu-etal-2021-vision} & 32.05 & 13.20 & 26.89 & 8.95 & 16.14 \\
LLaVA-v1.6-7B \cite{li2024llava} & 35.43 & 12.42 & 26.99 & 8.26 & 15.74 \\
Yi-VL-6B \cite{young2024yi} & 10.05 & 2.76 & 7.54 & 3.78 & 5.03 \\
Qwen2.5-VL-7B \cite{wang2024qwen2} & 34.63 & 13.76 & 28.87 & 8.65 & 18.58 \\
GPT-4o & 42.23 & 18.62 & 35.79 & 10.46 & 22.07 \\
\bottomrule
\end{tabular*}
\endgroup
\end{table}

\subsection{Main Results}
\label{sec:results}

Tables~\ref{tab:msairs} and~\ref{tab:mcdscs} report the main results. On MSAIRS, S2TDR improves fine-tuned text encoders, zero-shot LLMs, and the multimodal MMSAIR model over their original inputs. Compared with EI and ITEI, S2TDR gives more stable gains, especially for intent recognition, where implicit sticker acts need to be converted into explicit dialogue cues. This suggests that intent prediction benefits from pragmatic reconstruction rather than standalone sticker descriptions.

On MCDSCS, S2TDR generally improves summarization models across ROUGE, BLEU, and METEOR metrics. Although a few individual metrics are better with EI or ITEI, reconstruction-based integration yields the strongest overall trend across backbones. This is consistent with the nature of summarization: sticker information should be integrated into the discourse rather than appended as isolated visual annotations.

\subsection{Component Ablation}
\label{sec:ablation}

We ablate the main components of S2TDR on MMSAIR for MSAIRS and LLaMA2 for MCDSCS. Removing visual cues, affective-intent inference, or dialogue reconstruction consistently weakens performance. The visual and affective-intent components provide perceptual and pragmatic evidence, while dialogue reconstruction determines how these signals are expressed as a usable utterance.

Among the three components, removing dialogue reconstruction causes the clearest drop. This shows that obtaining sticker descriptions or emotion-intent labels alone is not sufficient; the key step is converting them into a coherent turn that fits the dialogue. We report ablations by removed component, which better reflects how each variant changes the reconstruction pipeline.

\subsection{Diagnostic Analysis}
\label{sec:analysis}

\paragraph{OCR-empty stickers}
Table~\ref{tab:ocr-coverage} shows that OCR text is unavailable for a substantial portion of stickers. In MSAIRS, 52.47\% of samples have empty sticker text; in MCDSCS, 32.30\% of unique stickers have empty OCR text. S2TDR therefore treats OCR as an auxiliary cue rather than a required input. For OCR-empty stickers, reconstruction relies on visual descriptions and dialogue context.

\paragraph{Reconstruction quality}
We further examine reconstruction quality through preference-based evaluation. As shown in Table~\ref{tab:quality}, S2TDR receives the highest preference share on faithfulness, fluency, and overall quality under both human and GPT-4 judgments. These values are preference shares rather than absolute quality scores. We therefore use this table as auxiliary diagnostic evidence rather than as a standalone human-evaluation benchmark.

\begin{table}[t]
\centering
\caption{Preference-based reconstruction quality.}
\label{tab:quality}
\begingroup
\footnotesize
\setlength{\tabcolsep}{6pt}
\renewcommand{\arraystretch}{0.88}
\begin{tabular}{llccc}
\toprule
Eval. & Criterion & S2TDR & VEI & LLaVA \\
\midrule
Human & Faithfulness & 0.66 & 0.19 & 0.15 \\
Human & Fluency & 0.51 & 0.35 & 0.14 \\
Human & Overall & 0.69 & 0.20 & 0.11 \\
\midrule
GPT-4 & Faithfulness & 0.68 & 0.14 & 0.18 \\
GPT-4 & Fluency & 0.55 & 0.30 & 0.15 \\
GPT-4 & Overall & 0.72 & 0.12 & 0.16 \\
\bottomrule
\end{tabular}
\endgroup
\end{table}

\paragraph{Case study}
Table~\ref{tab:cases} presents representative cases from both MSAIRS and MCDSCS. The cases show that S2TDR can recover useful sticker semantics for OCR-empty stickers and make sticker-derived affect explicit for both classification and summarization. They also reveal a key limitation: when the context is short or underspecified, reconstruction may add plausible but over-specific affective details.

\begin{table}[t]
\centering
\caption{Qualitative case study.}
\label{tab:cases}
\begingroup
\footnotesize
\renewcommand{\arraystretch}{1.08}
\setlength{\tabcolsep}{4pt}
\begin{tabularx}{0.96\linewidth}{@{}>{\raggedright\arraybackslash}p{0.18\linewidth}YYY@{}}
\toprule
Case & Context / summary clue & Sticker evidence & S2TDR effect \\
\midrule
MSAIRS: OCR-empty success
& Steam registration help request.
& OCR-empty sticker.
& Generates a fatigue-aligned reply for classification. \\
\addlinespace[2pt]
MSAIRS: ambiguous context
& Generic greeting with weak context.
& OCR-empty sticker.
& Adds a specific activity, showing over-specific reconstruction risk. \\
\addlinespace[2pt]
MCDSCS: summary support
& Singing praise; the summary requires positive affect.
& Two OCR-empty stickers.
& Converts stickers into happy and thankful turns. \\
\addlinespace[2pt]
MCDSCS: risk case
& Award display; the summary focuses on pride and humility.
& Two OCR-empty stickers.
& Adds confusion and surprise, showing over-interpretation risk. \\
\bottomrule
\end{tabularx}
\endgroup
\end{table}

\paragraph{Failure cases}
The risk cases suggest that S2TDR is not an error-free semantic parser. Ambiguous context, indirect sticker text, and noisy OCR may lead to over-specific reconstructions. Nevertheless, the intermediate visual, affective-intent, and reconstructed outputs make such errors easier to inspect than opaque multimodal predictions.

\subsection{Discussion and Limitations}
\label{sec:discussion}

S2TDR is designed as a reconstruction module rather than a replacement for multimodal dialogue models. Its main advantage is that sticker semantics are converted into an interpretable textual form, allowing existing text encoders, sequence-to-sequence models, and LLMs to use sticker information without changing their architectures. The intermediate outputs also make errors easier to inspect: visual descriptions, affective-intent summaries, and reconstructed utterances can be examined separately.

The framework still has limitations. First, reconstruction quality depends on the visual and language models used in the S2T process, so stronger or domain-adapted MLLMs may further improve performance. Second, converting stickers into text may lose fine-grained visual style, especially for highly stylized or culturally specific stickers. Third, S2TDR may over-specify the speaker's affect when the dialogue context is short or ambiguous. Future work can address these issues through uncertainty estimation, retrieval of similar sticker usages, and human-in-the-loop verification for high-risk cases.

\FloatBarrier
\section{Conclusion}
\label{sec:conclusion}

We studied sticker-aware dialogue understanding from the perspective of structured multimodal reconstruction. Instead of treating stickers as standalone images or auxiliary labels, S2TDR converts sticker turns into context-conditioned textual utterances through visual cue extraction, affective-intent inference, and dialogue reconstruction. Experiments on sentiment analysis, intent recognition, and dialogue summarization show that this reconstruction-based formulation improves multiple downstream models over direct sticker integration methods. Diagnostic analyses further show that OCR is only an auxiliary cue, component removals weaken performance, and dialogue-style reconstructions are generally preferred over description-style alternatives. These findings suggest that verbalizing the pragmatic function of stickers is an effective and interpretable way to bridge stickers and text-based dialogue models.

\bibliographystyle{elsarticle-num}
\bibliography{ccl2026}

\end{document}